\documentclass[final]{cvpr}

\usepackage{times}
\usepackage{epsfig}
\usepackage{graphicx}
\usepackage{amsmath}
\usepackage{amssymb}

\usepackage{bm}
\makeatletter
\@namedef{ver@everyshi.sty}{}

\DeclareRobustCommand\onedot{\futurelet\@let@token\@onedot}
\def\@onedot{\ifx\@let@token.\else.\null\fi\xspace}

\def\etal{\emph{et al}\onedot}

\makeatother
\usepackage{tikz}

\usepackage[percent]{overpic}

\usepackage[pagebackref=true,breaklinks=true,colorlinks,bookmarks=false]{hyperref}

\def\cvprPaperID{2973} 
\def\confYear{CVPR 2021}

\begin{document}

\title{Unsupervised Scale-Invariant Multispectral Shape Matching}

\author{Idan Pazi\\
Tel Aviv University\\
{\tt\small idan.pazi@outlook.com}

\and
Dvir Ginzburg\\
Tel Aviv University\\
{\tt\small dvirginzburg@mail.tau.ac.il}

\and
Dan Raviv\\
Tel Aviv University\\
{\tt\small darav@tauex.tau.ac.il}
}

\maketitle

\begin{abstract}
Alignment between non-rigid stretchable structures is one of the most challenging tasks in computer vision, as the invariant properties are hard to define, and there is no labeled data for real datasets. We present unsupervised neural network architecture based upon the spectral domain of scale-invariant geometry. We build on top of the functional maps architecture, but show that learning local features, as done until now, is not enough once the isometry assumption breaks. We demonstrate the use of multiple scale-invariant geometries for solving this problem. Our method is agnostic to local-scale deformations and shows superior performance for matching shapes from different domains when compared to existing spectral state-of-the-art solutions. 
\end{abstract}

\section{Introduction}
Shape correspondence is the problem of aligning features between two shapes and providing a pointwise map that reveals the underlying common anatomy of the shapes. The correspondence problem is especially challenging in the non-rigid case where stretching and bending are allowed, as shape features may differ substantially. The solution to this problem and its derivatives is vital for many applications, such as texture transfer, pose transfer, and space-time reconstruction.
\cite{survey_2011, survey_2012, num_geom_non_rigid}.
Research in the field of shape correspondence has found the general non-rigid correspondence problem to be incredibly tricky, as stable invariant properties are non-trivial to define. 
Trying to solve this problem directly, matching every point on the spatial domain yields a vast solution space, non-convex, and highly non-linear \cite{gh_dist_shapes}. The complexity of non-rigid shape correspondence problems led the efforts in this field to try and simplify the problem, primarily by exploiting invariants between shapes \cite{recent_trends}, focusing on metric-based methods where isometry or near-isometry is assumed between shapes \cite{gmds, isometry_invariant_pcl}
The success of metric-based approaches, in comparison to previous efforts, is based on the fundamental understanding that the features should be extracted from the manifold itself and not from the ambient space. 
Kernel methods such as the Heat Kernel Signature \cite{hks} and the Wave Kernel Signature \cite{wks} were highly useful; these descriptors rely on the intrinsic geometry of the shapes and are inherently invariant to isometries. Prominent \emph{Local Reference} methods analyze descriptors from the perspective of small patches on the shape and thus invariant to rotations\cite{local_surface_patch}. On each reference patch, a descriptive and unique signature can be extracted using hand-crafted algorithms such as accumulating vertices normals \cite{shot} or curvatures \cite{meshhog}. Recent works show a significant improvement of descriptors by applying machine learning tools for refining given descriptors \cite{fmnet} and using extracted features as descriptors \cite{geofmnet}. On the general case, finding appropriate descriptors for detecting signatures that are invariant to non-rigid deformations is not an easy task.

\begin{figure}
\centering
\begin{overpic}[width=0.9\linewidth]{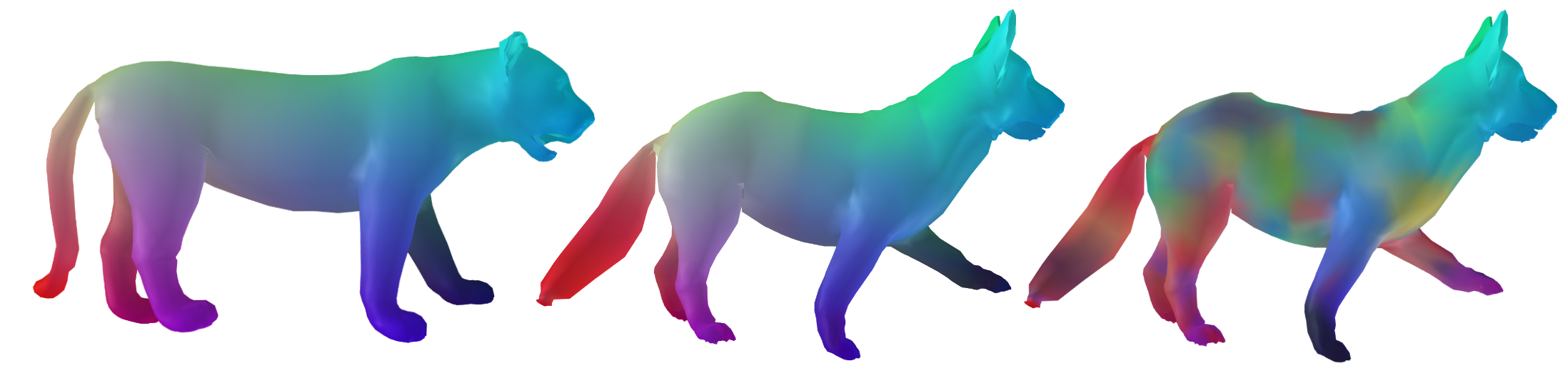}
    \put (16, 24) {Ref.}
    \put (49, 24) {Ours}
    \put (78, 24) {SURFMnet}
    \put (58, 1.6) {\fontsize{7pt}{0}$\mathbf{0.023}$}
    \put (93.5, 1.6) {\fontsize{7pt}{0}$0.168$}
\end{overpic}
\begin{overpic}[width=0.96\linewidth]{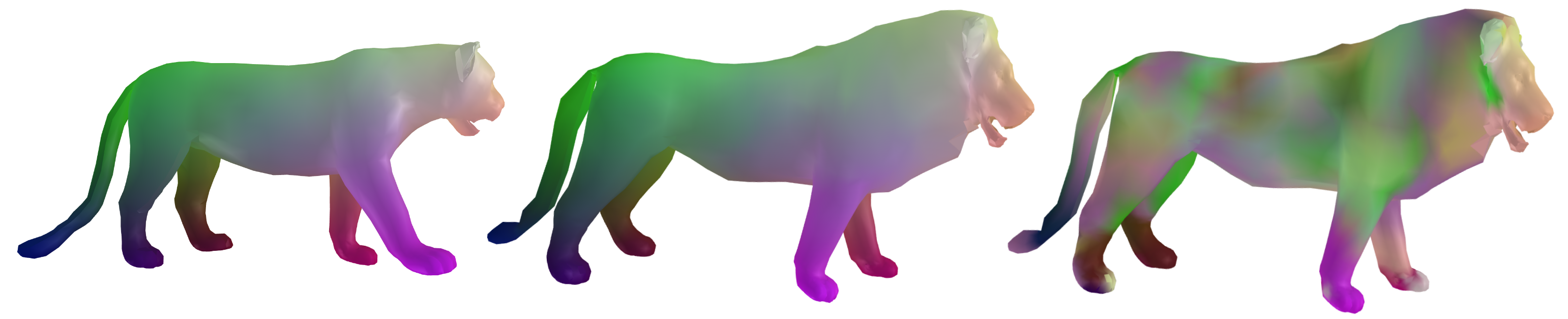}
    \put (58, 1.2) {\fontsize{7pt}{0}$\mathbf{0.005}$}
    \put (91.5, 1.2) {\fontsize{7pt}{0}$0.201$}
\end{overpic}
\caption{Unsupervised functional maps correspondence between highly non-isometric shapes from the SMAL dataset. Visualization of texture transfer. Mean geodesic error is marked next to each correspondence.}
\label{fig:smal}
\end{figure}
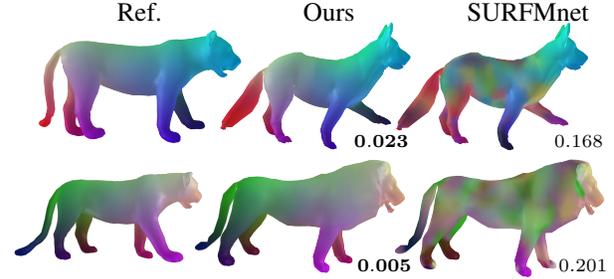

\emph{Spectral domain} methods for shape correspondence had increased popularity in recent years, commonly using the Laplace-Beltrami operator (LBO) eigendecomposition for spanning the spectral domain. 
The functional maps framework \cite{functional_maps}, showed it is possible to find a linear transformation between the spectral domains of different shapes and laid the foundations of recent advancements. Functional maps efficiently optimize the descriptors correspondence problem, shifting the problem from estimating similarity between descriptors on the complex spatial domain to solving a least-squares problem on a compact spectral domain.
Later works show how functional maps could be utilized in deep-learning architectures and outperform other methods on standard shape correspondence benchmarks \cite{fmnet, unsup_fmnet, surfmnet, geofmnet}.
In practice, the optimized linear functional map between LBO decompositions is sufficiently descriptive only between near-isometric shapes. Recent spectral methods for solving the correspondence problem are concentrated on matching near-isometric shapes with similar local and global properties \cite{fmnet, surfmnet, geofmnet}. Only limited attempts were made to adjust axiomatically the spectral correspondence mechanism for dealing with non-isometric shapes. Cyclic distortion \cite{cyclic} neglects the need for a consistent metric between shapes and successfully correspond stretchable domains but is still unable to correspond shapes from different domains. 

Aflalo \etal presented in \cite{si} a scale-invariant metric that is mathematically consistent under global scaling and showed that the LBO on this geometry is invariant to local scaling. One can argue that non-isometric shapes differ by a set of affine transformations and can be approximated by such a finite set; after inducing a scale-invariant metric, shapes can be treated as near-isometric, and a distance preserving map between the shapes can be obtained when measured by the alternative pseudo-metric.

\emph{In functional-based methods described above, the basis functions were derived using a Euclidean Laplace Beltrami Operator which varies significantly during non-isometric deformations, and tried to cope with that by learning local features per point.}

\subsection*{Contributions}

\begin{itemize} \itemsep-0.3em 
    \item We introduce a new architecture for non-rigid model alignment based on multispectral scale-invariant strategy, relaxing the near-isometry assumption.
    \item Provide superior alignment results, by a large margin, on scenarios where extensive stretching exists and open new possibilities to match non-isometric shapes using the functional maps framework.
    \item We report improved alignment results also on the near-isometry non-rigid benchmarks.
\end{itemize}
\section{Background}
\subsection{Riemannian Manifold}
A Riemannian manifold $(\mathcal{X}, g)$  in the 3-dimensional space is a parameterized surface  ${\mathcal{X}: \Omega_\mathcal{X} 	\subset \mathbb{R}^2 \rightarrow \mathbb{R}^3}$  with $g$ being a metric that varies differentiably with $p\in \Omega_\mathcal{X} $. The metric $g$  can be viewed as a family of inner products  ${g}_p(p_1, p_2) = \langle p_1, p_2 \rangle_p$ between vectors $p_1, p_2$ in the neighborhood of $p$. 
A \emph{map} $T: \Omega_\mathcal{X} \rightarrow  \Omega_\mathcal{Y}$ from one Riemannian manifold $\mathcal{X}$ to another manifold $\mathcal{Y}$  is a function that maps every point from one shape to the other.
Two Riemannian manifolds are \emph{isometric} if there is a map between them that preserves distances. 
\subsection{Laplace-Beltrami operator spectral domain}
The \emph{Laplace-Beltrami operator} (LBO) of $f: \Omega_\mathcal{X} \rightarrow \mathbb{R}$, a differentiable function over a manifold, is defined as:
\begin{equation} \label{eq:lbo}
\Delta_{g}f = -\frac{1}{\sqrt{\mathrm{det}\,g}}\sum_{ij} \frac{\partial}{\partial x_i}  ( g^{-1}\sqrt{\mathrm{det}\,g}\,\frac{\partial}{\partial x_j} f)  
\end{equation}
with $x_1,x_2$ are coordinates in $\Omega_\mathcal{X}$ and $g$ is the metric tensor ${g =\begin{pmatrix} g_{11} & g_{12} \\ g_{21} & g_{22} \end{pmatrix} , g_{ij} = \langle p_i, p_j \rangle_p}$.
 Hence, the operator is entirely defined by the metric tensor $g$.
It is a positive semi-definite operator and admits an eigendecomposition $\Delta_{g} \phi_i = \lambda_i \phi_i$ of an orthonormal basis ${\langle \phi_i,  \phi_j \rangle  = \delta_{ij}}$ and a function on the manifold can be presented as a Fourier series $f = \sum \langle \phi_i , f \rangle_\mathcal{X} \phi_i$.
It has been shown that the first $k$ eigenfunctions of the LBO decomposition are the optimal $k$-dimensional basis for representing smooth functions on $\mathcal{X}$ \cite{lbo_optimality}.

\subsection{Scale-invariant geometry}
Aflalo \etal \cite{si} constructed a metric on a parameterized differentiable surface that is invariant to scaling. Such metric $\tilde{g}$ could be achieved by adjusting the Euclidean metric $g$ by the magnitude of the Gaussian curvature $K$:
\begin{equation} \label{eq:pure_si_metric}
    \tilde{g} =|K|\,g
\end{equation}

With this pseudo metric, distances on the surface $\mathcal{X}$ are preserved when locally scaled by some factor. 
This pseudo-metric can be used to define a scale-invariant LBO $\Delta_{\tilde{g}}$ by substituting Eq. (\ref{eq:pure_si_metric}) into Eq. (\ref{eq:lbo}).

\subsection{Functional maps}

Functional maps \cite{functional_maps,fmap_course}, are spectral maps between pairs of shapes that allow efficient inference and manipulation. Given two parameterized manifolds ${\mathcal{X}: \Omega_\mathcal{X} \subset \mathbb{R}^2 \rightarrow \mathbb{R}^3}$ and ${\mathcal{Y}: \Omega_\mathcal{Y} \subset \mathbb{R}^2 \rightarrow \mathbb{R}^3}$, each with a space of real valued functions on the domain, $\mathcal{F}(\Omega_\mathcal{X},\mathbb{R})$ and $\mathcal{F}(\Omega_\mathcal{Y},\mathbb{R})$, our goal is to represent a bijective mapping between the two manifolds, namely ${T:\Omega_\mathcal{X} \rightarrow \Omega_\mathcal{Y}}$. 
Given some real-valued function ${f:\Omega_\mathcal{X}\rightarrow \mathbb{R}}$ on the first manifold, we may apply the mapping to obtain a corresponding function on the second manifold by ${f\circ 
T^{-1}: \Omega_\mathcal{Y}\rightarrow \mathbb{R}}$. This function composition can be viewed as a function transformation, represented by ${T_F:\mathcal{F}(\Omega_\mathcal{X},\mathbb{R}) \rightarrow  \mathcal{F}(\Omega_\mathcal{Y},\mathbb{R})}$. The goal mapping $T$ can be recovered from the function transformation $T_F$ by applying $T_F$ on a set of indicator functions with a single point support for every point in the domain. 
As opposed to the complex non-linear mapping $T$, the functional transformation $T_F$ is a linear map between function spaces. 
Assuming the function spaces $\mathcal{F}(\Omega_\mathcal{X},\mathbb{R})$ and  $\mathcal{F}(\Omega_\mathcal{Y},\mathbb{R})$ have orthonormal basis $\Phi$ and $\Psi$ respectively, any function $f:\Omega_\mathcal{X}\rightarrow \mathbb{R}$ can be represented as a linear combination $f=\sum_{i}{a_i \phi_i}$, and $T_F$ is a linear transformation:

\begin{equation} \label{eq:fm_c}
T_F(f)\!=\!T_F\left(\sum_{i}a_i \phi_i\right)\!=\!\sum_{i}a_i T_F(\phi_i )\!=\!\sum_{i}\sum_{j}a_i c_{ij} \psi_j
\end{equation}
Where $c_{ij}$ is the j'{th} coefficient in the basis span by $\Psi$ of the transformation of the i'{th} basis function, hence ${T_F(\phi_i)=\sum_{j}{c_{ij} \psi_j}}$, meaning that $T_F(\phi_i)$ determined only by the basis and the map $T$. On orthonormal basis functions we have $c_{ij} = \langle T_F(\phi_i), \psi_j \rangle$. By representing $c_{ij}$ as a matrix $C=(c_{ij})$ and $f$ as a vector of coefficients ${\vec{a}=(a_0,a_1,...,a_i,...)}$ the transformation takes the form of:
\begin{equation}
\label{eq:fm_c_matrix}
T_F(\vec{a}) = C\vec{a}
\end{equation}
The problem of matching descriptors of two shapes can be transformed from non-convex highly non-linear constraints about every point in each shape to a linear least-squares problem over the elements of the matrix $C$. Given a set of descriptors (real valued functions) on each shape $F$ and $G$ and their projection to the spectral domains - $F_\Phi$ and $G_\Psi$, finding $C_{\Phi\Psi}$, a linear transformation from the spectral domain $\Phi$ to $\Psi$, is a least-squares problem:
\begin{equation} \label{eq:func_map_opt}
	C_{\Phi\Psi}=\underset{C}{\mathrm{argmin}} ||CF_\Phi - G_\Psi||^2_F
\end{equation}
$C$ of significantly smaller size compared to the number of vertices can successfully describe the full correspondence with a low error when the LBO is used to span the spectral domain on each shape \cite{functional_maps,partial_functional_corr,functional_map_shape_collections,sparse_modeling}.
So far, the research in the field of functional maps is mainly focused on the standard LBO decomposition with the Euclidean metric as the basis of choice. The choice of basis is a fundamental component for the solution, as each shape descriptor is projected into it, and the capacity of $C$ to efficiently describe complex structures is derived from the strength of the space spanned by the basis.
In this work, the limitations of this choice are researched and relaxed by using an alternative scale-invariant LBO. 

\subsection{Deep functional maps}
Litany \etal made a paradigm shift in FMNet \cite{fmnet}, by training a supervised neural network for solving the shape correspondence problem without solving a labeling problem for each point but by refining baseline SHOT\cite{shot} descriptors and matching them using the functional maps framework. A soft error loss term was introduced for training the network, estimating the geodesic distance error at each point for a given functional map $C$ using the ground truth correspondence and the geodesic distances between points on each shape.

In SURFMNet \cite{surfmnet}, an unsupervised and more axiomatic approach for learning descriptors is suggested, where a set of numerical properties of the mapping matrix $C$ has been shown to have a strong correlation to superb alignment results in terms of geodesic error. Not only this architecture is fully unsupervised and reduces the need for expensive point labeling, but it is also completely independent of the geodesic distances on the shapes, which makes this architecture compelling for corresponding non-isometric shapes where the geodesic distance between corresponding points may differ substantially between shapes.
A different approach was suggested by Ginzburg and Raviv \cite{cyclic}, which focused on a cyclic architecture, where the distortion was measured only in the source domain. The cyclic approach provided resistance to stretching while also being fully unsupervised. Although each architecture was more robust, none of them were able to correspond shapes from different domains.

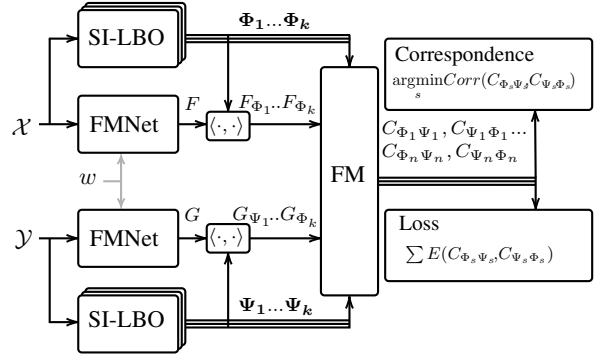
\begin{figure}

\tikzset{every picture/.style={line width=0.75pt}} 

\begin{tikzpicture}[x=0.75pt,y=0.75pt,yscale=-1,xscale=1,thick,scale=0.9, every node/.style={scale=0.9}]

\draw    (18,80) -- (38,80) ;
\draw [shift={(40,80)}, rotate = 180] [color={rgb, 255:red, 0; green, 0; blue, 0 }  ][line width=0.75]    (6.56,-1.97) .. controls (4.17,-0.84) and (1.99,-0.18) .. (0,0) .. controls (1.99,0.18) and (4.17,0.84) .. (6.56,1.97)   ;
\draw    (23.5,32) -- (38,32) ;
\draw [shift={(40,32)}, rotate = 180] [color={rgb, 255:red, 0; green, 0; blue, 0 }  ][line width=0.75]    (6.56,-1.97) .. controls (4.17,-0.84) and (1.99,-0.18) .. (0,0) .. controls (1.99,0.18) and (4.17,0.84) .. (6.56,1.97)   ;
\draw [color={rgb, 255:red, 178; green, 178; blue, 178 }  ,draw opacity=1 ]   (64,98) -- (64,126) ;
\draw [shift={(64,126)}, rotate = 270] [color={rgb, 255:red, 178; green, 178; blue, 178 }  ,draw opacity=1 ][line width=0.75]    (4.37,-1.96) .. controls (2.78,-0.92) and (1.32,-0.27) .. (0,0) .. controls (1.32,0.27) and (2.78,0.92) .. (4.37,1.96)   ;
\draw [shift={(64,98)}, rotate = 90] [color={rgb, 255:red, 178; green, 178; blue, 178 }  ,draw opacity=1 ][line width=0.75]    (4.37,-1.96) .. controls (2.78,-0.92) and (1.32,-0.27) .. (0,0) .. controls (1.32,0.27) and (2.78,0.92) .. (4.37,1.96)   ;
\draw [color={rgb, 255:red, 178; green, 178; blue, 178 }  ,draw opacity=1 ]   (54,112) -- (64,112) ;
\draw    (24,32) -- (24,80) ;
\draw    (192.48,32) -- (98,32) ;
\draw    (96,80) -- (110,80) ;
\draw [shift={(112,80)}, rotate = 180] [line width=0.75]    (6.56,-1.97) .. controls (4.17,-0.84) and (1.99,-0.18) .. (0,0) .. controls (1.99,0.18) and (4.17,0.84) .. (6.56,1.97)   ;
\draw    (136,80) -- (174,80) ;
\draw [shift={(176,80)}, rotate = 180] [line width=0.75]    (6.56,-1.97) .. controls (4.17,-0.84) and (1.99,-0.18) .. (0,0) .. controls (1.99,0.18) and (4.17,0.84) .. (6.56,1.97)   ;
\draw    (124,30) -- (124,71) ;
\draw [shift={(124,73)}, rotate = 270] [line width=0.75]    (6.56,-1.97) .. controls (4.17,-0.84) and (1.99,-0.18) .. (0,0) .. controls (1.99,0.18) and (4.17,0.84) .. (6.56,1.97)   ;
\draw    (192,34) -- (96,34) ;
\draw    (192,30) -- (192,46) ;
\draw [shift={(192,48)}, rotate = 270] [line width=0.75]    (6.56,-1.97) .. controls (4.17,-0.84) and (1.99,-0.18) .. (0,0) .. controls (1.99,0.18) and (4.17,0.84) .. (6.56,1.97)   ;
\draw    (208,112) -- (296.33,112) ;
\draw    (208,114) -- (296.33,114.01) ;
\draw    (24,191) -- (38,191) ;
\draw [shift={(40,191)}, rotate = 180] [color={rgb, 255:red, 0; green, 0; blue, 0 }  ][line width=0.75]    (6.56,-1.97) .. controls (4.17,-0.84) and (1.99,-0.18) .. (0,0) .. controls (1.99,0.18) and (4.17,0.84) .. (6.56,1.97)   ;
\draw    (24,144) -- (24,191.5) ;
\draw    (192,192) -- (97.52,192) ;
\draw    (124,194) -- (124,154) ;
\draw [shift={(124,152)}, rotate = 450] [color={rgb, 255:red, 0; green, 0; blue, 0 }  ][line width=0.75]    (6.56,-1.97) .. controls (4.17,-0.84) and (1.99,-0.18) .. (0,0) .. controls (1.99,0.18) and (4.17,0.84) .. (6.56,1.97)   ;
\draw    (18,144) -- (38,144) ;
\draw [shift={(40,144)}, rotate = 180] [color={rgb, 255:red, 0; green, 0; blue, 0 }  ][line width=0.75]    (6.56,-1.97) .. controls (4.17,-0.84) and (1.99,-0.18) .. (0,0) .. controls (1.99,0.18) and (4.17,0.84) .. (6.56,1.97)   ;
\draw    (96,144) -- (110,144) ;
\draw [shift={(112,144)}, rotate = 180] [color={rgb, 255:red, 0; green, 0; blue, 0 }  ][line width=0.75]    (6.56,-1.97) .. controls (4.17,-0.84) and (1.99,-0.18) .. (0,0) .. controls (1.99,0.18) and (4.17,0.84) .. (6.56,1.97)   ;
\draw    (136,144) -- (174,144) ;
\draw [shift={(176,144)}, rotate = 180] [color={rgb, 255:red, 0; green, 0; blue, 0 }  ][line width=0.75]    (6.56,-1.97) .. controls (4.17,-0.84) and (1.99,-0.18) .. (0,0) .. controls (1.99,0.18) and (4.17,0.84) .. (6.56,1.97)   ;
\draw    (96,194) -- (192,194) -- (191.98,178.76) ;
\draw [shift={(191.97,176.76)}, rotate = 449.92] [color={rgb, 255:red, 0; green, 0; blue, 0 }  ][line width=0.75]    (6.56,-1.97) .. controls (4.17,-0.84) and (1.99,-0.18) .. (0,0) .. controls (1.99,0.18) and (4.17,0.84) .. (6.56,1.97)   ;
\draw    (191.97,190) -- (100.24,190) ;
\draw    (296.33,110) -- (296.02,74) ;
\draw [shift={(296,72)}, rotate = 449.5] [line width=0.75]    (6.56,-1.97) .. controls (4.17,-0.84) and (1.99,-0.18) .. (0,0) .. controls (1.99,0.18) and (4.17,0.84) .. (6.56,1.97)   ;
\draw    (208,110) -- (296.33,110) -- (296.33,126.01) ;
\draw [shift={(296.33,128.01)}, rotate = 270] [color={rgb, 255:red, 0; green, 0; blue, 0 }  ][line width=0.75]    (6.56,-1.97) .. controls (4.17,-0.84) and (1.99,-0.18) .. (0,0) .. controls (1.99,0.18) and (4.17,0.84) .. (6.56,1.97)   ;
\draw    (192.37,30) -- (100.4,30) ;

\draw  [fill={rgb, 255:red, 255; green, 255; blue, 255 }  ,fill opacity=1 ] (44,14) .. controls (44,12.9) and (44.9,12) .. (46,12) -- (98,12) .. controls (99.1,12) and (100,12.9) .. (100,14) -- (100,42) .. controls (100,43.1) and (99.1,44) .. (98,44) -- (46,44) .. controls (44.9,44) and (44,43.1) .. (44,42) -- cycle ;
\draw  [fill={rgb, 255:red, 255; green, 255; blue, 255 }  ,fill opacity=1 ] (42,16) .. controls (42,14.9) and (42.9,14) .. (44,14) -- (96,14) .. controls (97.1,14) and (98,14.9) .. (98,16) -- (98,44) .. controls (98,45.1) and (97.1,46) .. (96,46) -- (44,46) .. controls (42.9,46) and (42,45.1) .. (42,44) -- cycle ;
\draw  [fill={rgb, 255:red, 255; green, 255; blue, 255 }  ,fill opacity=1 ] (40,66) .. controls (40,64.9) and (40.9,64) .. (42,64) -- (94,64) .. controls (95.1,64) and (96,64.9) .. (96,66) -- (96,94) .. controls (96,95.1) and (95.1,96) .. (94,96) -- (42,96) .. controls (40.9,96) and (40,95.1) .. (40,94) -- cycle ;
\draw  [fill={rgb, 255:red, 255; green, 255; blue, 255 }  ,fill opacity=1 ] (40,18) .. controls (40,16.9) and (40.9,16) .. (42,16) -- (94,16) .. controls (95.1,16) and (96,16.9) .. (96,18) -- (96,46) .. controls (96,47.1) and (95.1,48) .. (94,48) -- (42,48) .. controls (40.9,48) and (40,47.1) .. (40,46) -- cycle ;
\draw  [fill={rgb, 255:red, 255; green, 255; blue, 255 }  ,fill opacity=1 ] (112,75) .. controls (112,73.9) and (112.9,73) .. (114,73) -- (134,73) .. controls (135.1,73) and (136,73.9) .. (136,75) -- (136,87) .. controls (136,88.1) and (135.1,89) .. (134,89) -- (114,89) .. controls (112.9,89) and (112,88.1) .. (112,87) -- cycle ;
\draw  [fill={rgb, 255:red, 255; green, 255; blue, 255 }  ,fill opacity=1 ] (212,130) .. controls (212,128.9) and (212.9,128) .. (214,128) -- (326,128) .. controls (327.1,128) and (328,128.9) .. (328,130) -- (328,166) .. controls (328,167.1) and (327.1,168) .. (326,168) -- (214,168) .. controls (212.9,168) and (212,167.1) .. (212,166) -- cycle ;
\draw  [fill={rgb, 255:red, 255; green, 255; blue, 255 }  ,fill opacity=1 ] (176,50) .. controls (176,48.9) and (176.9,48) .. (178,48) -- (206,48) .. controls (207.1,48) and (208,48.9) .. (208,50) -- (208,174) .. controls (208,175.1) and (207.1,176) .. (206,176) -- (178,176) .. controls (176.9,176) and (176,175.1) .. (176,174) -- cycle ;
\draw  [fill={rgb, 255:red, 255; green, 255; blue, 255 }  ,fill opacity=1 ] (44,174) .. controls (44,172.9) and (44.9,172) .. (46,172) -- (98,172) .. controls (99.1,172) and (100,172.9) .. (100,174) -- (100,202) .. controls (100,203.1) and (99.1,204) .. (98,204) -- (46,204) .. controls (44.9,204) and (44,203.1) .. (44,202) -- cycle ;
\draw  [fill={rgb, 255:red, 255; green, 255; blue, 255 }  ,fill opacity=1 ] (42,176) .. controls (42,174.9) and (42.9,174) .. (44,174) -- (96,174) .. controls (97.1,174) and (98,174.9) .. (98,176) -- (98,204) .. controls (98,205.1) and (97.1,206) .. (96,206) -- (44,206) .. controls (42.9,206) and (42,205.1) .. (42,204) -- cycle ;
\draw  [fill={rgb, 255:red, 255; green, 255; blue, 255 }  ,fill opacity=1 ] (40,178) .. controls (40,176.9) and (40.9,176) .. (42,176) -- (94,176) .. controls (95.1,176) and (96,176.9) .. (96,178) -- (96,206) .. controls (96,207.1) and (95.1,208) .. (94,208) -- (42,208) .. controls (40.9,208) and (40,207.1) .. (40,206) -- cycle ;
\draw   [fill={rgb, 255:red, 255; green, 255; blue, 255 }  ,fill opacity=1 ] (40,130) .. controls (40,128.9) and (40.9,128) .. (42,128) -- (94,128) .. controls (95.1,128) and (96,128.9) .. (96,130) -- (96,158) .. controls (96,159.1) and (95.1,160) .. (94,160) -- (42,160) .. controls (40.9,160) and (40,159.1) .. (40,158) -- cycle ;
\draw  [fill={rgb, 255:red, 255; green, 255; blue, 255 }  ,fill opacity=1 ] (112,138) .. controls (112,136.9) and (112.9,136) .. (114,136) -- (134,136) .. controls (135.1,136) and (136,136.9) .. (136,138) -- (136,150) .. controls (136,151.1) and (135.1,152) .. (134,152) -- (114,152) .. controls (112.9,152) and (112,151.1) .. (112,150) -- cycle ;
\draw [fill={rgb, 255:red, 255; green, 255; blue, 255 }  ,fill opacity=1 ] (212,34) .. controls (212,32.9) and (212.9,32) .. (214,32) -- (326,32) .. controls (327.1,32) and (328,32.9) .. (328,34) -- (328,70) .. controls (328,71.1) and (327.1,72) .. (326,72) -- (214,72) .. controls (212.9,72) and (212,71.1) .. (212,70) -- cycle ;

\draw (3,137.4) node [anchor=north west][inner sep=0.75pt]    {$\mathcal{Y}$};
\draw (45,74) node [anchor=north west][inner sep=0.75pt]   [align=left] {FMNet};
\draw (1,75.4) node [anchor=north west][inner sep=0.75pt]    {$\mathcal{X}$};
\draw (111.5,74) node [anchor=north west][inner sep=0.75pt]  [font=\footnotesize] [align=center] {$\displaystyle \langle \cdot ,\cdot \rangle $};
\draw (44,25) node [anchor=north west][inner sep=0.75pt]   [align=left] {SI-LBO};
\draw (128,16) node [anchor=north west][inner sep=0.75pt]  [font=\footnotesize] [align=left] {$\boldsymbol{\Phi_{1}} ...\boldsymbol{\Phi _{k}}$};
\draw (39,106.4) node [anchor=north west][inner sep=0.75pt]    {$w$};
\draw (98,63) node [anchor=north west][inner sep=0.75pt]  [font=\footnotesize]  {$\mathnormal{F}$};
\draw (128,62) node [anchor=north west][inner sep=0.75pt]  [font=\footnotesize]  {$F_{\Phi_1} \!..F_{\Phi_k}$};
\draw (245,89.75) node  [font=\footnotesize] [align=left] {\begin{minipage}[lt]{52.36000000000001pt}\setlength\topsep{0pt}
$
C_{\Phi_1 \Psi _{1}} ,C_{\Psi_1 \Phi _{1}} ...\\
C_{\Phi_n \Psi _{n}} ,C_{\Psi_n \Phi _{n}}
$
\end{minipage}};
\draw (180,102) node [anchor=north west][inner sep=0.75pt]   [align=left] {FM};
\draw (265,153.13) node  [font=\small] [align=left] [scale=0.8] {$\sum E( C_{\Phi_s \Psi _s}\! {,}C_{\Psi_s \Phi _s})$};
\draw (44,183) node [anchor=north west][inner sep=0.75pt]   [align=left] {SI-LBO};
\draw (128,176) node [anchor=north west][inner sep=0.75pt]  [font=\footnotesize] [align=left] {$\displaystyle \boldsymbol{\Psi _{1}} ...\boldsymbol{\Psi _{k}}$};
\draw (45,138) node [anchor=north west][inner sep=0.75pt]   [align=left] {FMNet};
\draw (111.5,136.3) node [anchor=north west][inner sep=0.75pt]  [font=\footnotesize] [align=center] {$\displaystyle \langle \cdot ,\cdot \rangle $};
\draw (98,125.5) node [anchor=north west][inner sep=0.75pt]  [font=\footnotesize]  {$G$};
\draw (125,125) node [anchor=north west][inner sep=0.75pt]  [font=\footnotesize]  {$G_{\Psi_1} \!..G_{\Phi_k}$};
\draw (216,35) node [anchor=north west][inner sep=0.75pt]  [font=\small] [align=left] {Correspondence};
\draw (218.33,130) node [anchor=north west][inner sep=0.75pt]  [font=\small] [align=left] {Loss};
\draw (215,50) node [anchor=north west][inner sep=0.75pt][font==\small][scale=0.7]{$\underset{s}{\arg\!\min}  Corr( C_{\Phi_s \!\Psi _s} \!\!{,}C_{\Psi_s \!\Phi _s}\!)$};

\end{tikzpicture}

\caption{\textbf{Multispectral deep functional maps architecture.} Given two manifolds $\mathcal{X}, \mathcal{Y}$, each with a set of spectral domains $\{\Phi_s\},\{\Psi_s\}$ and inferred descriptors $F, G$ from the FMNet architecture. Correspondence between the descriptors is computed on each spectral domain yielding a set of transformations between the spectral domains of each shape (on both directions) $\{C_{\Phi_s\Psi_s}\}, \{C_{\Psi_s\Phi_s}\}$. The loss accumulated over all transformations. 
} 

\label{fig:arch}

\end{figure}

\section{Method}
This work facilitates multispectral scale-invariant geometry for non-isometric shape correspondence, inspired by recent deep learning spectral-domain architectures. Section \ref{Discrete scale-invariant LBO} and \ref{Scale-invariant spectral domain} describe our implementation of the discrete scale-invariant LBO and how the multispectral basis is constructed. Section \ref{Pointwise correspondence} describes the multispectral fusion stage of the pipeline, where we gather correspondences from multiple spectral domains into a single spatial correspondence. 
The complete architecture of our solution is presented in Figure \ref{fig:arch}, our method is the first to propose a multispectral learnable approach for the dense non-rigid correspondence problem.\footnote{Implementation is available at \href{https://github.com/idanpa/SURFMNet-Multispectral}{github.com/idanpa/SURFMNet-Multispectral}}

\subsection{Descriptors} \label{Descriptors}
We have used the FMNet \cite{fmnet} deep learning paradigm for generating local descriptors on the shapes. In the preprocessing stage, the hand-crafted SHOT algorithm was used for generating the baseline descriptors, where 352 initial descriptors are generated per point for each shape. During training and inference, each set of descriptors is refined through a series of 7 fully connected residual layers\cite{resnet}, with shared weights between the two shapes.
To ease the comparison to previous methods, the descriptor refinement stage was left unchanged, and our architecture bears the same amount of learned weights as in \cite{unsup_fmnet,surfmnet}.
Nonetheless, our inferred descriptors are more informative  as a result of evaluating their performance on richer spectral domains when training.

\subsection{Scale-invariant spectral domain} \label{Scale-invariant spectral domain}
While scale-invariant LBO induces robustness to local deformations, experiments show that inducing scale-invariant geometry impairs the global properties of the decomposition. To overcome this problem, Aflalo \etal \cite{lbo_optimality} introduced $\alpha \in [0,1]$ exponent parameter to interpolate between the Euclidean metric ($\alpha=0$) to the scale-invariant pseudo-metric ($\alpha=1$), defining an intermediate scale-invariant geometry:

\begin{equation} \label{eq:si_metric}
    \tilde{g_\alpha}=|K|^\alpha\langle p_{i},p_{j}\rangle
\end{equation}

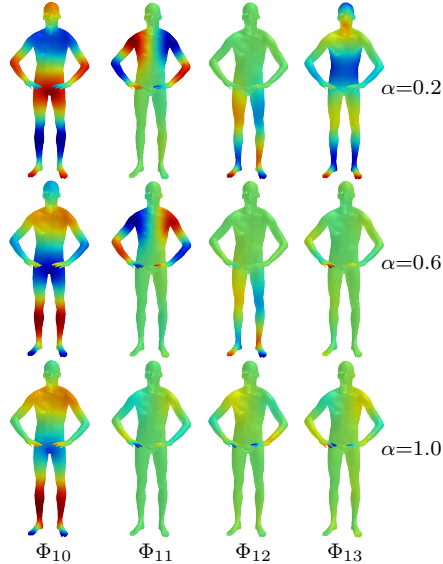
\begin{figure}
    \centering
    \begin{overpic}[width=0.6\linewidth]{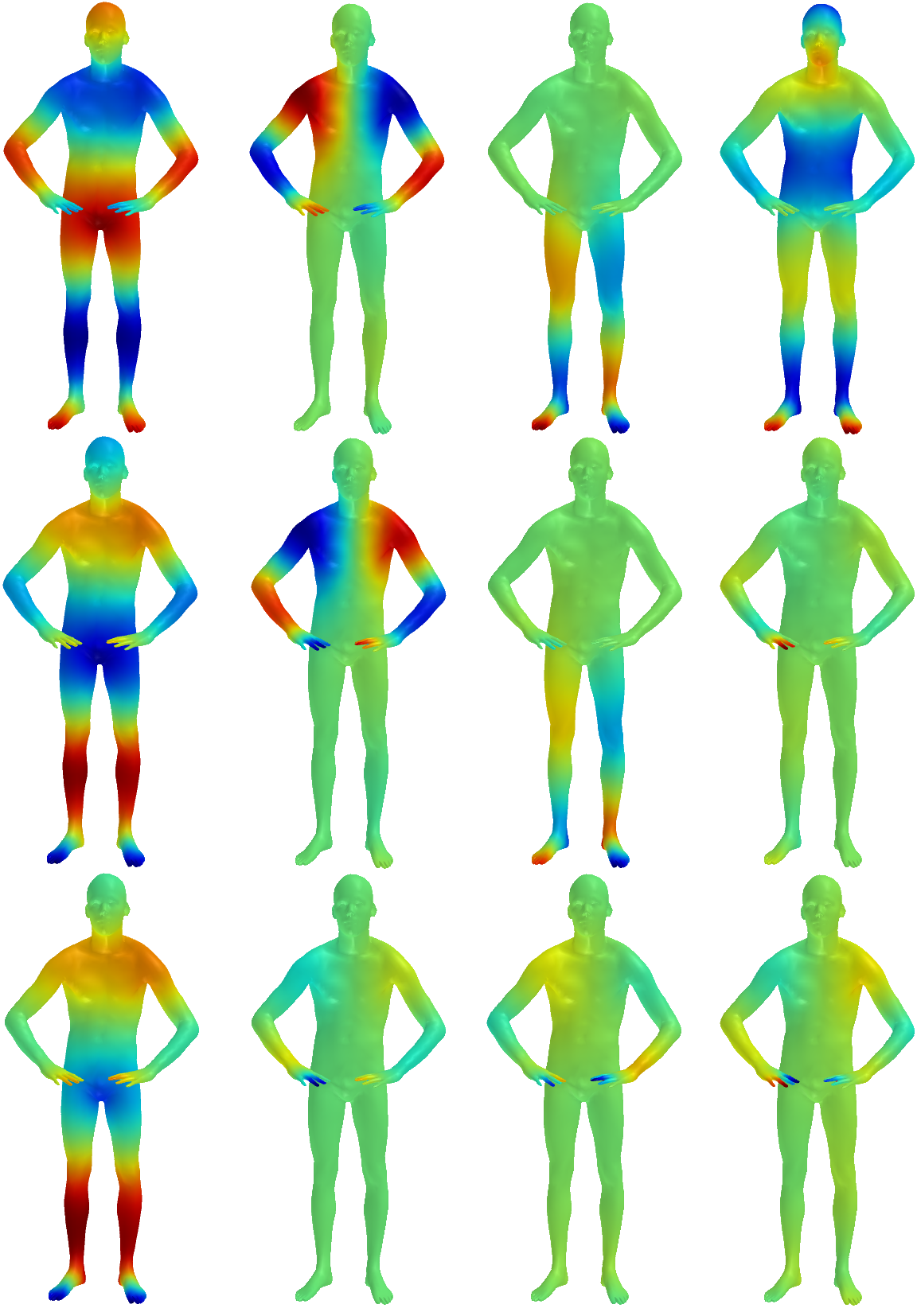}
        \put (5, -3) {\fontsize{8pt}{0}$\Phi_{10}$}
        \put (24, -3) {\fontsize{8pt}{0}$\Phi_{11}$}
        \put (42, -3) {\fontsize{8pt}{0}$\Phi_{12}$}
        \put (59, -3) {\fontsize{8pt}{0}$\Phi_{13}$}
        
        \put (69, 83.0) {\fontsize{8pt}{0}$\alpha{=}0.2$}
        \put (69, 50.5) {\fontsize{8pt}{0}$\alpha{=}0.6$}
        \put (69, 16.0) {\fontsize{8pt}{0}$\alpha{=}1.0$}
    \end{overpic}
    \vspace{10pt}
    \caption{The 10th to 13th eigenfunctions of scale-invariant LBO with different $\alpha$ values. On more scale-invariant metrics (higher $\alpha$), the eigenfunctions concentrates on local features.}
    \label{fig:alpha_cmp}
\end{figure}

A visualization of the impact of $\alpha$ on the spectral domain is shown in Figure \ref{fig:alpha_cmp}. 
The choice of $\alpha$ is an essential ingredient of the scale-invariant architecture presented in \cite{silbo_descr} where it was manually optimized for each pair of shapes. Instead of using a fixed $\alpha$ that would set a specific suboptimal trade-off between local and global features we choose a set of $\alpha$'s that we have found to be optimal for a wide range of shapes. 
We are the first to introduce a framework that infers correspondence by observing a set of spectral domains with different $\alpha$ values.

The Estimation of the Gaussian curvature on a discrete triangular mesh representation of a manifold is an active research topic \cite{gauss_curv_survey, gauss_curv_broken}. Our estimation of the Gaussian curvature was done by first applying Laplacian smoothing \cite{laplacian_smoothing}, then estimating the Gaussian curvature on the modified mesh according to the second fundamental form, following the algorithm presented in \cite{estimating_curv}. Curvature was clipped to a minimal value, to avoid vanishing distances on flat areas \cite{si}, and trimmed to a maximal value, to reduce the quantization noise of small triangles.

\subsection{Discrete scale-invariant LBO} \label{Discrete scale-invariant LBO}
The general approach for the discretization of the LBO on a triangulated surface with vertices ${\{\vec{v}_i\}}_{i=0}^{n-1}$  each with immediate neighbors ${\{N(i)\}}_{i=0}^{n-1}$ is to represent the operator as:
\begin{equation}
\Delta f (\vec{v_i}) := \frac{1}{d_i} \sum_{j\in N(i)} w_{ij}( f (\vec{v_i}) -  f (\vec{v_j}) )
\end{equation}
with some \emph{mass} $d_i$ associated to each vertex and a \emph{weight} $w_{ij}$ associated to each edge. 
We use the cotangent weights \cite{cot_lbo} and mass $d_i = a(i)/3$  where $a(i)$ is the scale-invariant area of all triangles at vertex $i$. Our implementation is based on the finite-element approach from \cite{reuter_lbo}, achieving numerical stability by formulating the problem as a generalized eigenvalues problem:
\begin{equation}
	A\vec{\phi} = -\lambda B \vec{\phi}
\end{equation}
\begin{equation}
    A(i,j) := \begin{cases}
      \frac{\operatorname{cot}(\alpha_{ij}) + \operatorname{cot}(\beta_{ij})}{2} & \text{$(i,j)$ edge}\\
      -\sum_{k \in N(i)} A(i,k) & i=j \\
      0 & \text{otherwise}
    \end{cases}  
\end{equation}

\begin{equation}
    B(i,j) := \begin{cases}
      \frac{|K_1|^\alpha|t_1|+|K_2|^\alpha|t_2|}{12} & \text{$(i,j)$ edge}\\
      \frac{\sum_{l \in N(i)} |K_l|^\alpha|t_l|}{6} & i=j \\
      0 & \text{otherwise}
    \end{cases}  
\end{equation}

$\alpha_{ij}$ and $\beta_{ij}$ denote the two angles opposite to the edge ${(i, j)}$, $|t|$ is the Euclidean area of the triangle $t$, $K_i$ is the Gaussian curvature associated to the triangle $t_i$. $t_1$, $t_2$ are the triangles sharing the edge ${(i, j)}$ and $\{t_l\}_{l \in N(i)}$ are the triangles in the immediate neighborhood of vertex $i$. $\alpha$ is the interpolation parameter from Eq. (\ref{eq:si_metric}). 

\subsection{Loss function} \label{Loss function}
Our loss function follows the unsupervised loss term presented in \cite{surfmnet}, where a set of regularization terms applied on the functional maps $C^i_{\Phi\Psi}$  and $C^i_{\Psi\Phi}$ from Eq. (\ref{eq:func_map_opt}).

Bijectivity of the transformation can be achieved by enforcing identity for the transformation from one spectral domain to the other and back:
\begin{equation}
E_{1} = ||C_{\Phi\Psi}C_{\Psi\Phi}-I||_2^2+||C_{\Psi\Phi}C_{\Phi\Psi}-I||_2^2
\end{equation}
Area preservation can be enforced by the orthogonality of each transformation:
\begin{equation}
E_{2} = ||C_{\Phi\Psi}^\top C_{\Phi\Psi}-I||_2^2+||C_{\Psi\Phi}^\top C_{\Psi\Phi}-I||_2^2
\end{equation}
A pointwise map expresses an intrinsic isometry if and only if the associated functional map commutes with the LBO\cite{functional_maps,lbo_on_manifold}:
\begin{equation}
E_{3} = ||C_{\Phi\Psi} \Lambda_\Phi -\Lambda_\Psi C_{\Phi\Psi} ||^2  +  ||C_{\Psi\Phi} \Lambda_\Psi -\Lambda_\Phi C_{\Psi\Phi} ||^2
\end{equation}
where $\Lambda_\Phi$ and $\Lambda_\Psi$ are diagonal matrices of the eigenvalues of $\Phi$ and $\Psi$ respectfully.

Functional map $T$ can represent a pointwise map if and only if it preserves the pointwise product $\bigodot$  between functions, namely $T(f\bigodot g) = T(f)\bigodot T(g) $ \cite{composition_operators_on_func_spaces}. Nogneng and Ovsjanikov followed this observation in \cite{surfmnet_desc_comm_loss} to present a penalty for descriptor preservation via commutativity:
\begin{multline}
\shoveleft{E_{4} = \hfill} \\ \sum_{f_i \in F_\Phi; g_i \in G_\Phi} \!\!\!\!\!\!\!\!\!\!\!||C_{\Phi\Psi} M_{f_i} - M_{g_i} C_{\Phi\Psi}||^2 + ||C_{\Psi\Phi} M_{g_i} - M_{f_i} C_{\Psi\Phi}||^2 \\ 
M_{f_i} = \Phi^+ Diag(f_i) \Phi, M_{g_i} = \Psi^+ Diag(g_i) \Psi 
\end{multline}

Where $F_\Phi$ and $G_\Psi$  are the shape descriptors with the spectral domains $\Phi$ and $\Psi$ respectfully, and $+$ is the Moore-Penrose pseudoinverse.

Given a set of functional maps $Sp = \{ (C_{\Phi_s\Psi_s}, C_{\Psi_s\Phi_s}) \}_{s=1}^k$ for each of the spectral domain pairs $\Phi_s \Psi_s$. The loss function accumulates loss over all spectral domains:
\begin{equation}
E = \sum_{C_{\Phi_s\Psi_s}, C_{\Psi_s\Phi_s} \in Sp} \quad \sum_{i \in \{1,2,3,4\}} \omega_i E_i (C_{\Phi_s\Psi_s},C_{\Psi_s\Phi_s})
\end{equation}
with heuristic weights $\omega_i$ from \cite{surfmnet}, where this loss term has been shown to strongly correlate to the desired low geodesic error of the pointwise correspondence. 

The loss function is differentiable with respect to the weights and depends only on the functional maps themselves \cite{surfmnet}, allowing unsupervised network training without a need for the ground truth correspondence, measuring expensive geodesic distances on shapes, or even inferring the pointwise correspondence while training. Instead of solving the non-convex assignment problem over thousands of distances between each point, we solve the least-squares problem over small spectral matrices. 
Unlike other works that presented similar penalties on the functional maps \cite{functional_maps,map_based_exploration,coupled_func_maps}, the structural penalty on the functional maps is decoupled from the descriptor similarity optimization of Eq. (\ref{eq:func_map_opt}).

\begin{figure}
    \centering
    \begin{overpic}[width=1\linewidth]{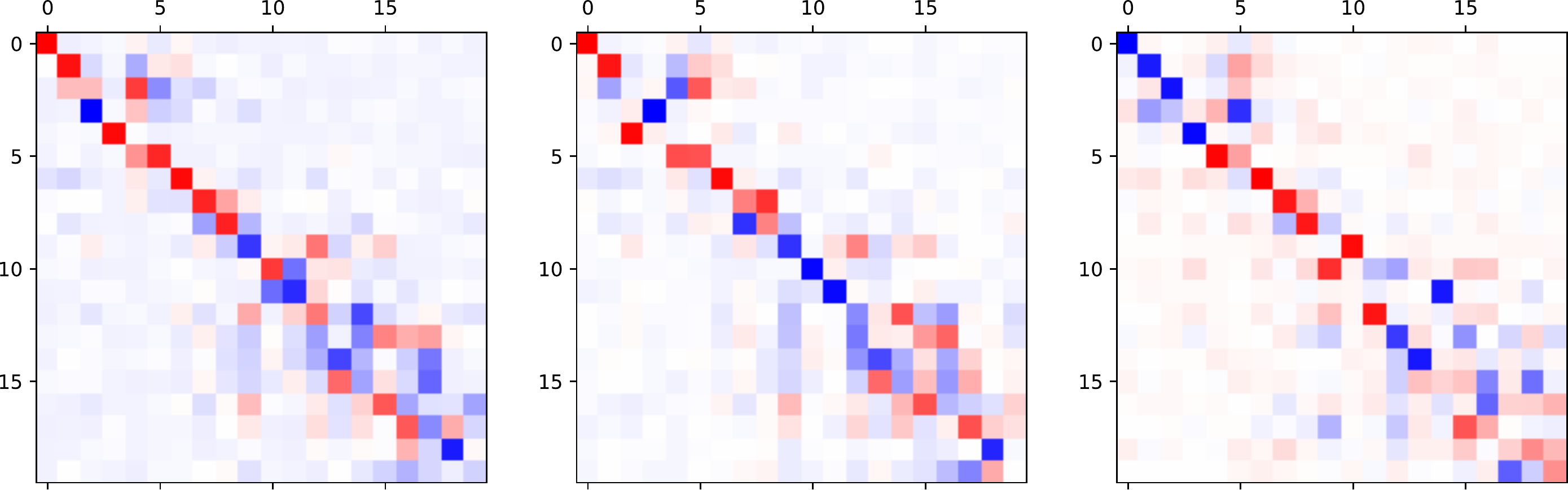}
        \put (5,  -4) {\fontsize{8pt}{0}\textbf{(a)} $C_{(\alpha=0.5)}$}
        \put (39, -4) {\fontsize{8pt}{0}\textbf{(b)} $C_{(\alpha=0.6)}$}
        \put (74, -4) {\fontsize{8pt}{0}\textbf{(c)} $C_{(\alpha=0.8)}$}
    \end{overpic}\\
    \vspace{8pt}
    \begin{overpic}[width=0.7\linewidth]{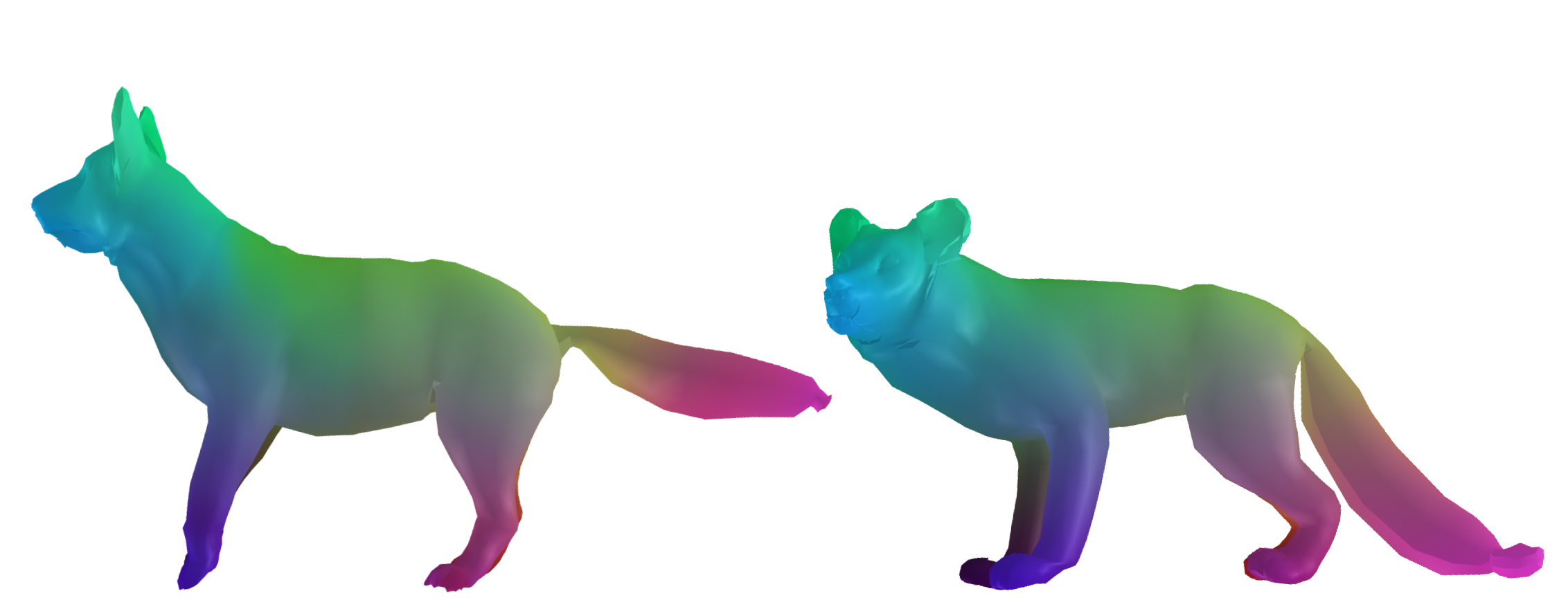}
        \put (48,  -2) {\fontsize{8pt}{0}\textbf{(d)}}
    \end{overpic}
    \vspace{8pt}
    \caption{Multispectral functional maps fusion. (a)(b)(c) are the functional maps for the different intermediate scale-invariant geometry (showing the first 30 eigenvectors). (d) is the combined correspondence from a dog (right) to a wolf (left) on the spatial domain.}
    \label{fig:c_fusion}
\end{figure}

\subsection{Pointwise correspondence from multispectral functional maps} \label{Pointwise correspondence}
Obtaining the pointwise correspondence between shapes from the functional map $C$ is not trivial \cite{pointwise_from_map}.
In \cite{functional_maps}, an efficient method to find the pointwise correspondence is presented, based upon proximity search in the spectral domain. Given a spectral domain for each shape represented by matrices $\Phi$ on $\mathcal{X}$ and $\Psi$ on $\mathcal{Y}$, where each column corresponds to a point and each row to an eigenfunction, for vertex $i$ on $\mathcal{Y}$, the corresponding point on $\mathcal{X}$ is the nearest neighbor:
\begin{equation}
	Corr(i) = \underset{j}{\mathrm{argmin}}||col_j(C\Phi) - col_i(\Psi)||_2
\end{equation}
Where $Corr(i)$ is the index of the corresponding vertex to the i'{th} vertex and $col_i(M)$ is the i'{th} column of the matrix M. 
Given multiple spectral maps, we wish to blend the individual maps and infer a single pointwise correspondence. Direct proximity search within all spectral domains is not feasible as distances between domains are not comparable. To overcome this problem, the multispectral scheme finds the optimal correspondence by normalizing the mean and range of all distances in the different spectral domains and selects the best fit in terms of normalized distance on multiple pairs of spectral domains ${\Phi_s}, {\Psi_s}$ with functional map $C_s$ between them: 
\begin{equation} \label{eq:corr_argmin}
	Corr(i) = \underset{j}{\mathrm{argmin}} \, \underset{s}{\mathrm{min}}||col_j(C_s\Phi_s) - col_i(\Psi_s)||^*_2
\end{equation}
Where $||\cdot||^*_2$ is the $L_2$ distance with the following normalization, on each spectral-domain we linearly normalize the distances to $[0,1]$ and then subtract the mean of all the distances in the spectral domain, establishing a comparable distance between spectral domains. By fusing the correspondences, we can compensate for the degraded performance of scale-invariant metrics while withstanding non-isometric deformations. Figure \ref{fig:c_fusion} illustrates this stage.

\section{Experiments}
We have tested our method on different datasets, showing an improvement over recent state-of-the-art solutions and highlighting the strength of our framework to overcome challenging cross-domain problems yet to be addressed using spectral methods. 

For each shape, three different spectral domains were generated on the preprocessing stage, with heuristic intermediate metric factor $\alpha$. For the near-isometric datasets we have used ${\alpha = 0, 0.6, 0.8}$ and for the non-isometric datasets we have used $\alpha = 0.5, 0.6, 0.8$. The Gaussian curvature clipping was made to the 0.4\% - 75\% range on each shape.

Training was done for 10k iterations on FAUST dataset, for SMAL dataset training was done for another 10k iteration on top of the trained model, we have used a learning rate of 0.001 with an ADAM optimizer.

\begin{figure}
    \centering
    \scalebox{.7}{\input{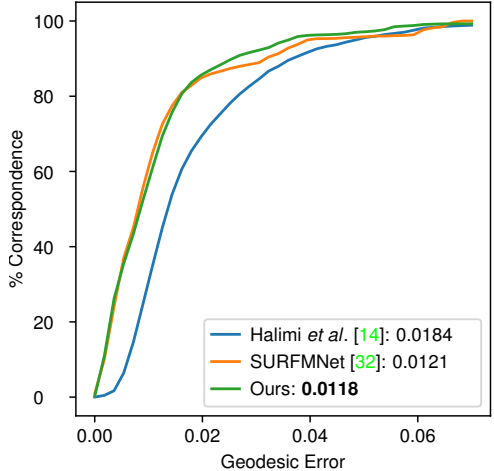}}
    \caption{Error evaluation on FAUST dataset compared to \cite{surfmnet} and \cite{unsup_fmnet} on inter-subject and intra-subject pairs. Our multispectral architecture shows improvement even on the near-isometric benchmark.}
    \label{fig:faust_graph}
\end{figure}
\subsection{Synthetic FAUST}
For evaluating our method on the near-isometry case, we compared our method to \cite{surfmnet} and \cite{unsup_fmnet} on the synthetic FAUST dataset \cite{faust}. This dataset contains synthetic scans of ten different human subjects, each on ten different poses. As suggested in \cite{fmnet}, we trained our model on the first eight subjects and tested it on the other two subjects. In Figure \ref{fig:faust_graph} we present the geodesic error comparison. Even in the near-isometry case, which our method was not optimized for, we manage to outperform state-of-the-art solutions with the same amount of weights. This shows how recent spectral domain methods bears an isometry assumption that not only limiting the capabilities to correspond shapes from different domains but also has an impact on the near-isometry case, where these methods are found to be sensitive to the stretching between different poses and different subjects.

\subsection{Locally scaled FAUST}
To highlight the ability of our method to tolerate deformations and the sensitivity of current methods to non-isometry cases, we have conducted the following experiment. We locally scaled examples from the FAUST dataset, and applied the scaling to the SHOT descriptors as well, simulating perfect scale-invariant descriptors. We have deformed the shapes by applying local uniform scaling of different parts of the bodies, such as head, arms, legs, and hands while keeping the shapes smooth as possible. The dataset preserved many of the characteristics of the original FAUST dataset and was designed to only stress the ability of spectral methods to handle local scaling while keeping most of the parameters optimized to the nature of the dataset. 

\begin{figure}
    \centering
    \begin{overpic}[width=0.98\linewidth]{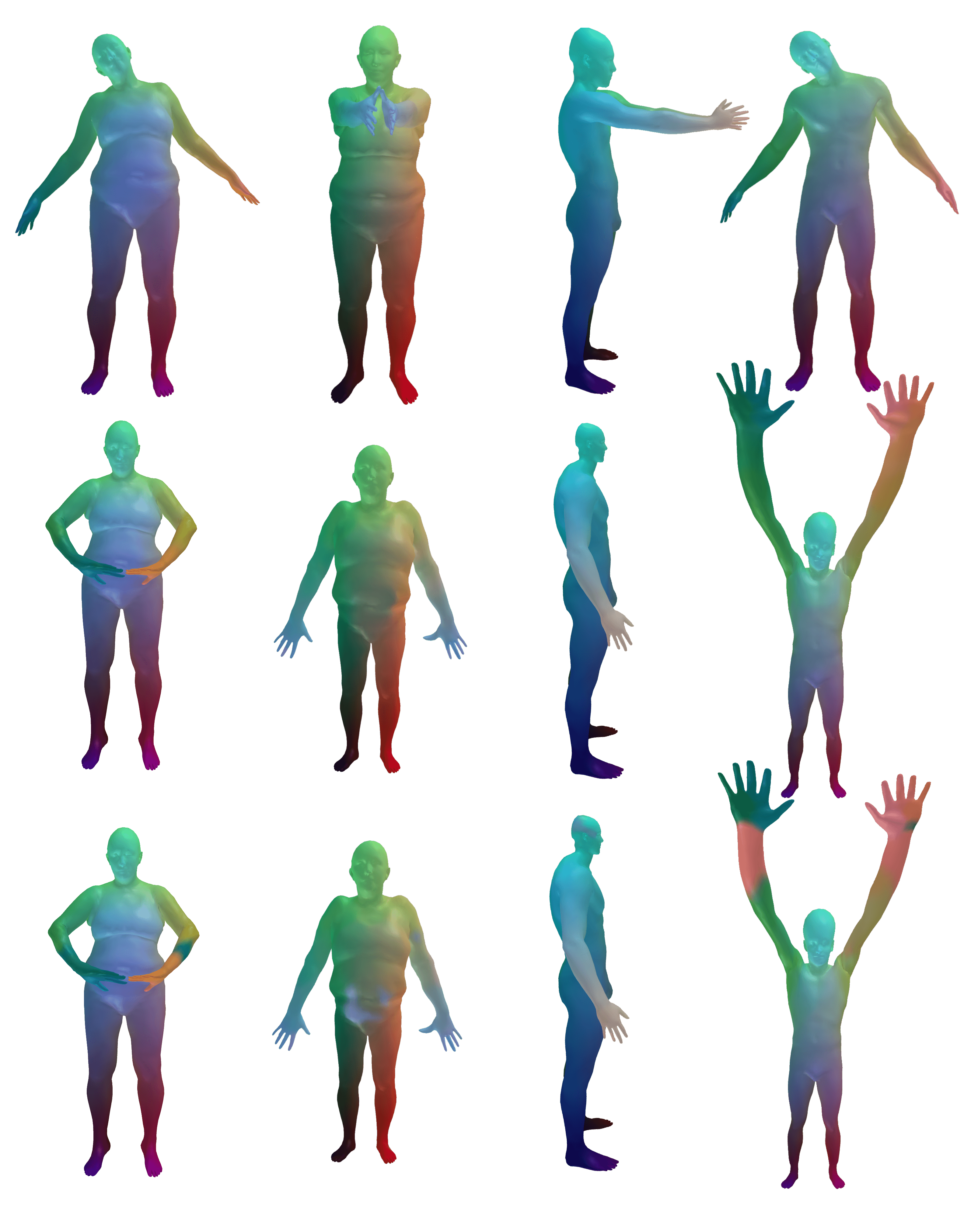}
    \end{overpic}
    \caption{Correspondence between shapes on deformed FAUST dataset, first row is the reference shapes, second row is the correspondence using scale-invariant LBO $\alpha{=}0.6$, third row is the correspondence using the Euclidean LBO. Scale-invariant geometry allows the architecture to be agnostic to deformations even without training on a deformed dataset.}
    \label{fig:faust_ls}
\end{figure}

\begin{figure}
    \centering
    \scalebox{.62}{\input{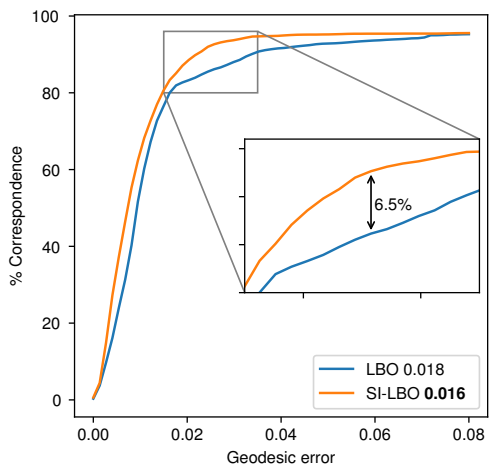}}
    \caption{Error evaluation on locally scaled FAUST dataset comparing the performance of the standard and scale-invariant LBO on SURFMNet\cite{surfmnet}}
    \label{fig:faust_ls_graph}
\end{figure}

We compared the performance of SURFMNet\cite{surfmnet} using the standard LBO and using a single scale-invariant spectral domain ($\alpha{=}0.6$) both trained only on the standard FAUST dataset. A visualization of the correspondence is presented in Figure \ref{fig:faust_ls}. The results show how scaling disturbs the matching capabilities and, in some cases, has a global impact on the correspondence of Euclidean-based approaches. Scale-invariant LBO excels on this dataset and even without training on locally scaled examples, managed to achieve almost the same performance as in the near-isometry case. We present a quantitative analysis of the geodesic errors on 32 different pairs from the dataset in Figure \ref{fig:faust_ls_graph}. This finding shows how current methods are surprisingly sensitive to minor local deformations on the standard near-isometry benchmark they were optimized for.  


\subsection{SMAL}
The most significant strength of our architecture is the ability to correspond shapes from different domains, an ability we are the first to implement using an unsupervised functional maps framework. We have evaluated our method on pairs of shapes from the SMAL dataset \cite{smal,smalr}. In Figure \ref{fig:smal} we present a visualization of the inter-class results. The results show how the relaxation of the near-isometry assumption gave the model the degree of freedom to generalize itself for corresponding non-isometric shapes. On the spectral domain of the scale-invariant geometry, our method managed to model inter-class relationships between shapes. We have observed how standard methods fail to converge in terms of the loss function and were giving completely broken matches for inter-class matching.

\section{Conclusions}
We have realized an end-to-end architecture for non-rigid and non-isometric shape correspondence, based upon scale-invariant geometry. Our architecture surpasses state-of-the-art methods on standard shape correspondence benchmarks, and it is the first to solve end-to-end inter-domain correspondence problems using the functional maps paradigm, where all other existing unsupervised spectral methods fail.
This work demonstrates the advantages of using multiple scale-invariant spectral domains for the task of shape correspondence. Without questioning the optimality of the LBO for spanning spectral domains, we show how the Euclidean metric is suboptimal for the purpose of non-rigid shape correspondence.
While presenting superb results on deformable pairs, challenging cases such as hippo and a cat can still be improved. We believe this work sets another step toward a fully general non-rigid shape correspondence method. The investigation of alternative metrics and bases should be further examined, such as the affine-invariant metric presented by Raviv \etal in \cite{affine} and the task-driven decomposition suggested by Azencot and Lai in \cite{pod}.

\section{Acknowledgments}
This work is partially funded by the Zimin Institute for Engineering Solutions Advancing Better Lives, the Israeli consortiums for soft robotics and autonomous driving, the Nicholas and Elizabeth Slezak Super Center for Cardiac Research and Biomedical Engineering at Tel Aviv University and TAU Science Data and AI Center.

\bibliographystyle{ieee_fullname}
\bibliography{egbib}

\end{document}